\pdfoutput=1
\documentclass{article}
\usepackage[final,nonatbib]{nips_2017}

\usepackage[utf8]{inputenc} 
\usepackage[T1]{fontenc}    
\usepackage{hyperref}       
\usepackage{url}            
\usepackage{booktabs}       
\usepackage{amsfonts}       
\usepackage{nicefrac}       
\usepackage{microtype}      

\usepackage{graphicx} 
\usepackage{subcaption} 
\usepackage{wrapfig}

\usepackage{algorithm}
\usepackage{algorithmic}
\usepackage{xr}

\usepackage{amsmath}
\usepackage{amssymb}
\usepackage{color}
\usepackage{mynotation} 

\title{Linearly constrained Gaussian processes}

\author{
	Carl Jidling\\
	Department of Information Technology\\
	Uppsala University, Sweden\\
	\texttt{carl.jidling@it.uu.se} \\
	\And
	Niklas Wahlstr\"{o}m \\
	Department of Information Technology\\
	Uppsala University, Sweden \\
	\texttt{niklas.wahlstrom@it.uu.se} \\
	\AND
	Adrian Wills \\
	School of Engineering \\
	University of Newcastle, Australia \\
	\texttt{adrian.wills@newcastle.edu.au} \\
	\And
	Thomas B. Sch\"{o}n \\
	Department of Information Technology\\
	Uppsala University, Sweden \\
	\texttt{thomas.schon@it.uu.se} \\
}

\begin{document}
	\newcommand{\coverTitle}{Linearly constrained Gaussian Processes}
	\newcommand{\coverAuthors}{Carl Jidling, Niklas Wahlstr{\"o}m, Adrian Wills and Thomas B. Sch{\"o}n}
	\newcommand{\coverYear}{2015}
	\newcommand{\coverStatus}{Accepted for publication.}
	
	\begin{titlepage}
		\begin{center}
			{\large \em Technical report}
			
			\vspace*{2.5cm}
			%
			{\Huge \bfseries \coverTitle  \\[0.4cm]}
			
			%
			{\Large \coverAuthors \\[2cm]}
			
			\renewcommand\labelitemi{\color{red}\large$\bullet$}
			\begin{itemize}
				\item {\Large \textbf{Please cite this version:}} \\[0.4cm]
				\large
				\coverAuthors. \coverTitle. \textit{Advances in Neural Information Processing Systems (NIPS)},
				Long Beach, CA, USA, December, 2017.	
			\end{itemize}

			\vfill
			
			\begin{abstract}
				We consider a modification of the covariance function in Gaussian processes to correctly account for known linear operator constraints.
				By modeling the target function as a transformation of an underlying function, the constraints are explicitly incorporated in the model such that they are guaranteed to be fulfilled by any sample drawn or prediction made.
				We also propose a constructive procedure for designing the transformation operator and illustrate the result on both simulated and real-data examples.
			\end{abstract}
			
			
			\vfill
		\end{center}
	\end{titlepage}

	\maketitle
	
	\begin{abstract} 
		We consider a modification of the covariance function in Gaussian processes to correctly account for known linear operator constraints.
		By modeling the target function as a transformation of an underlying function, the constraints are explicitly incorporated in the model such that they are guaranteed to be fulfilled by any sample drawn or prediction made.
		We also propose a constructive procedure for designing the transformation operator and illustrate the result on both simulated and real-data examples.
	\end{abstract}
	
	\begin{wrapfigure}{r}{0.5\textwidth}
		\vspace{-20pt}
		\begin{center}
			\centerline{\includegraphics[width=0.4\textwidth]{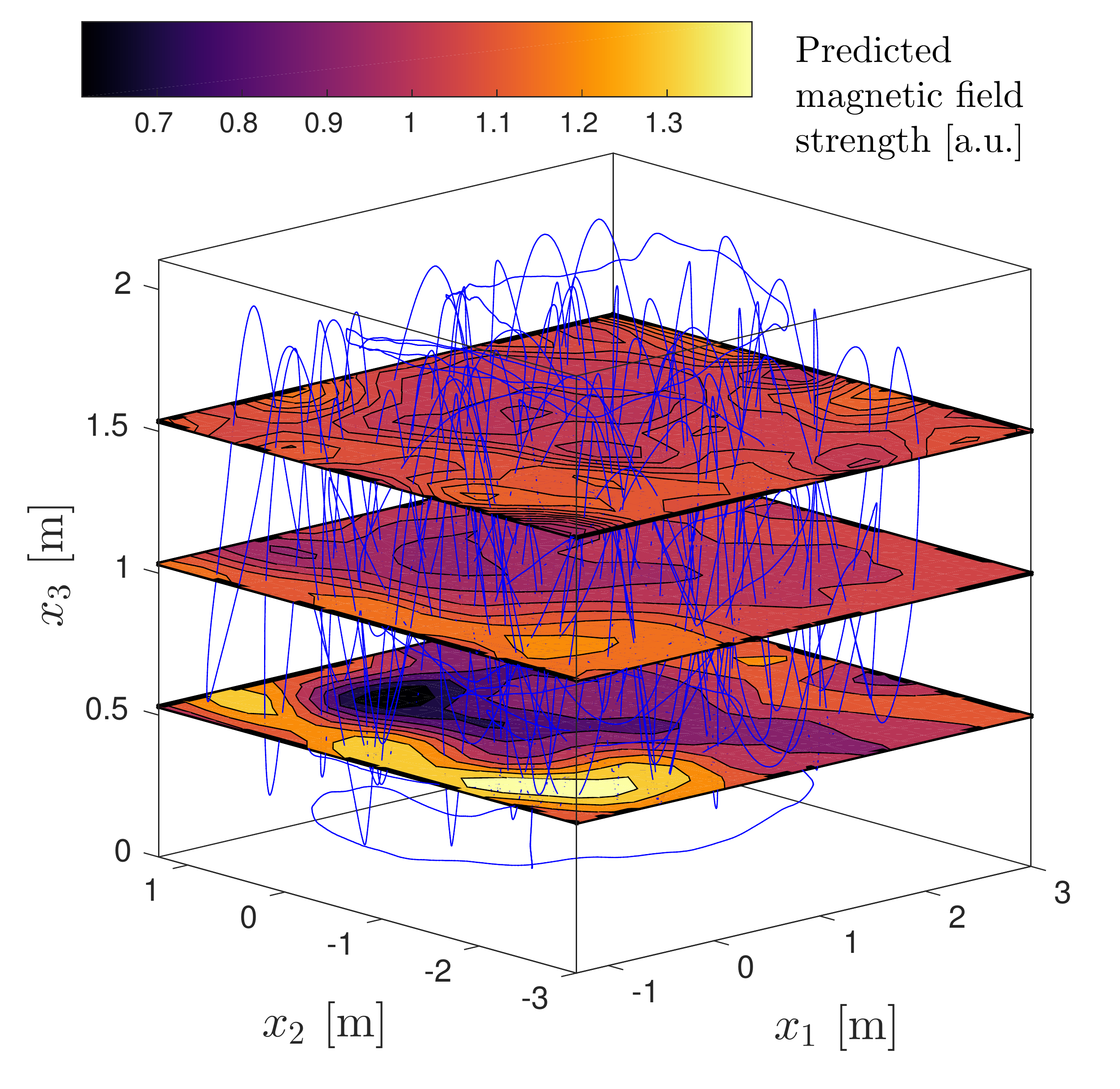}}
			\caption{Predicted strength of a magnetic field at three heights, given measured data sampled from the trajectory shown (blue curve). The three components $(x_1,x_2,x_3)$ denote the Cartesian coordinates, where the $x_3$-coordinate is the height above the floor. The magnetic field is curl-free, which can be formulated in terms of three linear constraints. The method proposed in this paper 
				can exploit these constraints to improve the predictions. See Section~\ref{sec:realDataExp} for details.} 
			\label{fig:sliceplot}
		\end{center}
		\vspace{-25pt}
	\end{wrapfigure} 
	
	\section{Introduction}\label{sec:intro}
	Bayesian non-parametric modeling has had a profound impact in machine
	learning due, in no small part, to the flexibility of these model structures in
	combination with the ability to encode prior knowledge in a principled
	manner~\cite{Zoubin2015}. 
	These properties have been exploited within the class of Bayesian
	non-parametric models known as Gaussian Processes (GPs), which have received significant research attention and have
	demonstrated utility across a very large range of real-world
	applications~\cite{Rasmussen2006}.

	Abstracting from the myriad number of these applications, it has been observed that the efficacy of GPs modeling is often intimately
	dependent on the appropriate choice of mean and covariance functions,
	and the appropriate tuning of their associated hyper-parameters.
	Often, the most appropriate mean and covariance functions are
	connected to prior knowledge of the underlying problem. 
	For example, \cite{Koyejo2013} uses functional expectation constraints
	to consider the problem of  gene-disease association, and
	\cite{Frellsen} employs a multivariate generalized	von Mises
	distribution to produce a GP-like regression that handles circular
	variable problems.
	
	At the same time, it is not always obvious how one might construct a GP model that obeys underlying principles, such as
	equilibrium conditions and conservation "laws". 
	One straightforward approach to this problem is to add fictitious
	measurements that observe 
	the constraints at a finite number of points of interest.
	This has the benefit of being relatively straightforward to implement, but has the sometimes significant 
	drawback of increasing the problem dimension and at the same time not enforcing the constraints between the points of interest.
	
	A different approach to constraining the GP model is to
	construct mean and covariance functions that obey the constraints.
	For example, curl and divergence free covariance functions are used in
	\cite{WahlstromPHD} 
	to improve the accuracy for regression
	problems. 
	The main benefit of this approach is that the problem
	dimension does not grow, and the constraints are enforced everywhere, not pointwise.  
	However, it is not obvious how
	these approaches can be scaled for an arbitrary set of
	linear operator constraints.

	The contribution of this paper is a new way to include constraints into multivariate GPs. 
	In particular, we develop a method that
	transforms a given GP into a new, derived, one that satisfies the constraints.
	The procedure relies upon the fact that GPs are closed under linear operators, and we propose an algorithm capable of constructing the required transformation.
	We will demonstrate the utility of this new method on both simulated
	examples and on a real-world application, the latter in form of predicting the components of a magnetic field, as illustrated in Figure~\ref{fig:sliceplot}.
	
	To make these ideas more concrete, we present a simple example that
	will serve as a focal point several times throughout the paper. 
	To that end, assume that we have a two-dimensional function $\f(\x):
	\mathbb{R}^2 \mapsto \mathbb{R}^2$ on which we put a GP prior
	$
	\f(\x) \sim \GPd{\vec{\mu}(\x)}{\K(\x,\x')}.
	$
	We further know that $\f(\x)$ should obey the differential equation
	\begin{align} \label{eq:intro_example}
	\pd{f_1}{x_1}+\pd{f_2}{x_2}=0.
	\end{align}
	In this paper we show how to
	modify $\K(\x,\x')$ and $\vec{\mu}(\x)$ such that any sample from the new GP is guaranteed to obey the constraints like \eqref{eq:intro_example}, considering any kind of linear operator constraint.

	\section{Problem formulation}
	Assume that we are given a data set of $N$ observations $\{\x_k, \y_k\}_{k=1}^N$ where $\x_k$ denotes the input and $\y_k$ the output. Both the input and output are potentially vector-valued, where $\x_k \in \mathbb{R}^D$ and $\y_k \in \mathbb{R}^K$.
	We consider the regression problem where the data can be
	described by a non-parametric model
	$
	\y_k = \f(\x_k) + \e_k,
	$
	where $\e_k$ is zero-mean white noise representing the measurement uncertainty.	
	In this work, we place a vector-valued GP prior on $\f$ 
	\begin{align} \label{eq:gp_prior}
	\f(\x) \sim \GPd{\vec{\mu}(\x)}{\K(\x,\x')},
	\end{align}	
	with the mean function and the covariance function
	\begin{align} 
	\label{eq:mean_coavaraince_function}
	\vec{\mu}(\cdot): \,\, \mathbb{R}^D \mapsto \mathbb{R}^K, \qquad \qquad \qquad
	\K(\cdot,\cdot): \,\, \mathbb{R}^D \times \mathbb{R}^D \mapsto \mathbb{R}^K \times \mathbb{R}^K.
	\end{align}
	Based on the data $\{\x_k, \y_k\}_{k=1}^N$, we would now like
	to find a posterior over the function $\f(\x)$. In addition to the data, we know that the function $\f$ should fulfill certain constraints
	\begin{align} \label{eq:constraints}
	\Fcalmat_\x[\f]=\vec{0},
	\end{align}
	where $\Fcalmat_\x$ is an operator mapping the function $\f(\x)$ to another function $\g(\x)$ as  $\Fcalmat_\x[\f] = \g(\x)$. We further require $\Fcalmat_\x$ to be a linear operator meaning that
	$
	\Fcalmat_\x\Big[\lambda_1\f_1+\lambda_2\f_2\Big] = \lambda_1\Fcalmat_\x[\f_1]+\lambda_2\Fcalmat_\x[\f_2],
	$
	where $\lambda_1, \, \lambda_2 \in \mathbb{R}$.	
	The operator $\Fcalmat_\x$ can for example be a linear transform
	$\Fcalmat_\x[\f] = \C \f(\x)$ which together with the
	constraint \eqref{eq:constraints} forces a certain linear
	combination of the outputs to be linearly dependent. 
	
	The operator $\Fcalmat_\x$ could also include other linear operations on the function $\f(\x)$.
	For example, we might know that the function $\f(\x):\mathbb{R}^2\rightarrow\mathbb{R}^2$ should obey a certain partial differential equation $\Fcalmat_\x[\f] = \pd{f_1}{x_1}+\pd{f_2}{x_2}$. 
	A few more linear operators are listed in Section~\ref{app:lin_op} of the Supplementary material, including integration as one the most well-known.
	
	The constraints \eqref{eq:constraints} can either come from known physical laws or other prior knowledge of the process generating the data. 
	Our objective is to encode these constraints in the mean and covariance functions \eqref{eq:mean_coavaraince_function} such that any sample from the corresponding GP prior \eqref{eq:gp_prior} always obeys the constraint \eqref{eq:constraints}.
	
	\section{Building a constrained Gaussian process}\label{sec:constrainedGP}
	
	\subsection{Approach based on artificial observations}\label{sec:naiveAppr}
	Just as Gaussian distributions are closed under linear transformations, so are GPs closed under linear operations (see Section~\ref{app:gp_lin_op} in the Supplementary material). 
	This can be used for a straightforward way of embedding linear operator constraints of the form \eqref{eq:constraints} into GP regression. 
	The idea is to treat the constraints as noise-free artificial observations $\{\tilde \x_k,\tilde \y_k\}_{k=1}^{\tilde N}$ with $\tilde \y_k = \vec{0}$ for all $k=1\dots\tilde N$. 
	The regression is then performed on the model
	$
	\tilde \y_k = \Fcalmat_{\tilde \x_k}[\f],
	$
	where $\tilde \x_k$ are input points in the domain of interest.		
	For example, one could let these artificial inputs $\tilde \x_k$ coincide with the points of prediction. 
	
	An advantage of this approach is that it allows constraints of the type \eqref{eq:constraints} with a non-zero right hand side.
	Furthermore, there is no theoretical limit on how many constraints we can include (i.e. number of rows in $\Fcalmat_\x$) -- although in practice, of course, there is. 
	
	However, this is problematic mainly for two reasons. 
	First of all, it makes the problem size grow.
	This increases memory requirements and execution time, and the numerical stability is worsen due to an increased condition number. 
	This is especially clear from the fact that we want these observations to be noise-free, since the noise usually has a regularizing effect.
	Secondly, the constraints are only enforced point-wise, so a sample drawn from the posterior fulfills the constraint only in our chosen points.
	The obvious way of compensating for this is by increasing the number of points in which the constraints are observed -- but that exacerbates the first problem.  
	Clearly, the challenge grows quickly with the dimension of the inferred function. 
	
	
	Embedding the constraints in the covariance function removes these issues -- it makes the enforcement continuous while the problem size is left unchanged.
	We will now address the question of how to design such a covariance function.
	
	\subsection{A new construction}	 			 	
	We want to find a GP prior \eqref{eq:gp_prior} such that any sample $\f(\x)$ from that prior obeys the constraints \eqref{eq:constraints}. In turn, this leads to constraints on the mean and covariance functions \eqref{eq:mean_coavaraince_function} of that prior. However, instead of posing these constraints on the mean and covariance functions directly, we consider $\f(\x)$ to be related to another function $\g(\x)$ via some operator $\Gcalmat_\x$
	\begin{align} \label{eq:mapping}
	\f(\x) = \Gcalmat_\x[\g].
	\end{align}
	The constraints \eqref{eq:constraints} then amounts to
	\begin{align} \label{eq:operator_cond1}
	\Fcalmat_\x[\Gcalmat_\x [\g]] = \mat{0}.
	\end{align} 						
	We would like this relation to be true for any function $\g(\x)$. 
	To do that, we will interpret $\Fcalmat_\x$ and $\Gcalmat_\x$
	as matrices and use a similar procedure to that of solving
	systems of linear equations. Since $\Fcalmat_\x$ and $\Gcalmat_\x$  are linear operators, we can think of $\Fcalmat_\x[\f]$ and $\Gcalmat_\x[\g]$ as matrix-vector multiplications where
	$
	\Fcalmat_\x[\f] = \Fcalmat_\x \f, \quad \text{with} \quad
	(\Fcalmat_\x \f)_i = \sum_{j=1}^K (\Fcal_\x)_{ij} f_j
	$
	where each element $(\Fcal_\x)_{ij}$ in the operator matrix $\Fcalmat_\x$ is a scalar operator. With this notation, \eqref{eq:operator_cond1} can be written as					
	\begin{align} \label{eq:operator_cond2}
	\Fcalmat_\x \Gcalmat_\x = \mat{0}.
	\end{align} 	
	
	This reformulation imposes constraints on the operator $\Gcalmat_\x$ rather than on the GP prior for $\f(\x)$ directly.
	We can now proceed by designing a GP prior for $\g(\x)$ and transform it using the mapping \eqref{eq:mapping}. We further know that GPs are closed under linear operations. More specifically, if $\g(\x)$ is modeled as a GP with mean $\vec{\mu}_\g(\x)$ and covariance $\K_\g(\x,\x')$,
	then $\f(\x)$ is also a GP with	
	\begin{align}\label{eq:f_from_g}
	\f(\x) = \Gcalmat_\x \g \sim \GPd{\Gcalmat_\x\ \vec{\mu_\g}}{\Gcalmat_\x \K_\g \Gcalmat_{\x'}^\Transp}.
	\end{align} 	
	We use $ (\Gcalmat_\x \K_\g
	\Gcalmat_{\x'}^\Transp)_{ij}$ to denote that
	$
	(\Gcalmat_\x \K_\g \Gcalmat_{\x'}^\Transp)_{ij} = (\Gcal_\x)_{ik} (\Gcal_{\x'})_{jl} (K_\g)_{kl},
	$
	where $\Gcalmat_{\x}$ and $\Gcalmat_{\x'}$ act on the first and second argument of $\K_\g(\x,\x')$, respectively. See Section~\ref{app:gp_lin_op} in the Supplementary material for further details on linear operations on GPs.
	
	The procedure to find the desired GP prior for $\f$ can now be divided into the following three steps
	\begin{enumerate}
		\itemsep-1mm 
		\item Find an operator $\Gcal_\x$ that fulfills the condition \eqref{eq:operator_cond1}. 
		\item Choose a mean and covariance function for $\g(\x)$.
		\item Find the mean and covariance functions for $\f(\x)$ according to \eqref{eq:f_from_g}.
	\end{enumerate}
	
	In addition to being resistant to the disadvantages of the approach described in Section \ref{sec:naiveAppr}, there are some additional strengths worth pointing out with this method. 
	First of all, we have separated the task of encoding the constraints and encoding other desired properties of the kernel.
	The constraints are encoded in $\Fcalmat_\x$ and the remaining properties are determined by the prior for $\g(\x)$, such as smoothness assumptions. 
	Hence, satisfying the constraints does not sacrifice any desired behavior of the target function. 
	
	Secondly, $\K(\x,\x')$ is guaranteed to be a valid covariance function provided that $\K_\g(\x,\x')$ is, since GPs are closed under linear functional transformations.
	From \eqref{eq:f_from_g}, it is clear that each column of $\K$ must fulfill all constraints encoded in $\Fcalmat_\x$. 
	Possibly $\K$ could be constructed only with this knowledge, assuming a general form and solving the resulting equation system. 
	However, a solution may not just be hard to find, but one must also make sure that it is indeed a valid covariance function. 
	
	Furthermore, this approach provides a simple and straightforward way of constructing the covariance function even if the constraints have a complicated form. 
	It makes no difference if the linear operators relate the components of the target function explicitly or implicitly -- the procedure remains the same. 
	
	\subsection{Illustrating example}\label{sec:2d_divfree}
	We will now illustrate the method using the example
	\eqref{eq:intro_example} introduced already in the
	introduction. 
	Consider a function $\f(\x):\mathbb{R}^2\mapsto\mathbb{R}^2$  satisfying	
	$
	\pd{f_1}{x_1}+\pd{f_2}{x_2}=0,
	$
	where $\x = [x_1,\,\, x_2]^\Transp$ and $\f(\x) = [f_1(\x),\,\, f_2(\x)]^\Transp$. 
	This equation describes all two-dimensional divergence-free vector fields. 
	The constraint can be written as a linear constraint on the form \eqref{eq:constraints} where
	$
	\Fcalmat_\x=
	[
	\pd{}{x_1} \,\, \pd{}{x_2}
	]
	$
	and
	$
	\f(\x)=	
	[
	f_1(\x) \,\, f_2(\x)
	]^\Transp
	.
	$
	Modeling this function with a GP and building the covariance structure as described above, we first need to find the transformation $\Gcalmat_\x$ such that \eqref{eq:operator_cond2} is fulfilled. For example, we could pick
	\begin{equation} \label{eq:Gx1}
	\Gcalmat_\x=
	\begin{bmatrix}
	-\pd{}{x_2} & \pd{}{x_1}
	\end{bmatrix}^\Transp.
	\end{equation}
	If the underlying function $g(\x):\mathbb{R}^2\mapsto\mathbb{R}$ is given by
	$
	g(\x)\sim\GP\big(0,k_g(\x,\x')\big),
	$
	then we can make use of \eqref{eq:f_from_g} to obtain
	$
	\f(\x) \sim\GP\big(\vec{0},\K(\x,\x')\big)
	$
	where
	\begin{align*}
	\K(\x,\x') =
	\Gcalmat_\x k_g(\x,\x') \Gcalmat_\x^\Transp 
	=
	\begin{bmatrix}§
	\pd{^2}{x_2x_2'} & -\pd{^2}{x_2x_1'} \\[2mm]
	-\pd{^2}{x_1x_2'} & \pd{^2}{x_1x_1'}
	\end{bmatrix}k_g(\x,\x').
	\end{align*}
	Using a covariance function with the following structure, we know that the constraint  will be fulfilled by any function generated from the corresponding GP.
	

	\section{Finding the operator \texorpdfstring{$\Gcalmat_\x$}{Gx}}
	\label{sec:parametric_approach}
	

	In a general setting it might be hard to find an operator $\Gcalmat_\x$ that fulfills the constraint \eqref{eq:operator_cond2}. 
	Ultimately, we want an algorithm that can construct $\Gcalmat_\x$ from a given $\Fcalmat_\x$.
	In more formal terms, the function $\Gcalmat_\x\g$ forms the nullspace of $\Fcalmat_\x$. 
	The concept of nullspaces for linear operators is well-established \cite{Luenberger1969}, and does in many ways relate to real-number linear algebra.
	
	However, an important difference is illustrated by considering a one-dimensional function $f(x)$ subject to the constraint $\Fcal_x f = 0$ where $\Fcal_x = \pd{}{x}$. 
	The solution to this differential equation can not be expressed in terms of an arbitrary underlying function, but it requires $f(x)$ to be constant. 
	Hence, the nullspace of $\pd{}{x}$ consists of the set of horizontal lines. 
	Compare this with the real number equation $ab=0$, $a\neq0$, which is true only if $b=0$. 
	Since the nullspace differs between operators, we must be careful when discussing the properties of $\Fcalmat_\x$ and $\Gcalmat_\x$ based on knowledge from real-number algebra.
	
	
	Let us denote the rows in $\Fcalmat_\x$ as $\fcalvec_1^\Transp,\dots,\fcalvec_L^\Transp$. We now want to find all solutions $\gcalvec$ such that
	\begin{align} \label{eq:constraints_vec}
	\Fcalmat_\x \gcalvec = \vec{0} \quad \Rightarrow \quad	\fcalvec_i^\Transp \gcalvec = 0, \quad \forall \quad i=1,\dots, L.
	\end{align}
	The solutions $\gcalvec_1,\dots,\gcalvec_P$ to \eqref{eq:constraints_vec} will then be the columns of $\Gcalmat_\x$. Each row vector $\fcalvec_j$ can be written as $\fcalvec_i = \mat{\Phi}_i \vec{\xi}^\fcalvec$ where $\mat{\Phi}_i \in \mathbb{R}^{K \times M_\fcalvec}$ and $\vec{\xi}^\fcalvec = [\xi_1, \dots, \xi_{M_\fcalvec}]^\Transp$ is a vector of $M_\fcalvec$ scalar operators included in $\Fcalmat_\x$. 
	We now assume that $\gcalvec$ also can be written in a similar form $\gcalvec = \mat{\Gamma} \vec{\xi}^\gcalvec$ where $\mat{\Gamma} \in \mathbb{R}^{K \times M_\gcalvec}$ and $\vec{\xi}^\gcalvec = [\xi_1, \dots, \xi_{M_\gcalvec}]^\Transp$ is a vector of $M_\gcalvec$ scalar operators. One may make the assumption that the same set of operators that are used to describe $\fcalvec_i$ also can be used to describe $\gcalvec$, i.e., $\vec{\xi}^\gcalvec = \vec{\xi}^\fcalvec$. 
	However, this assumption might need to be relaxed.
	The constraints \eqref{eq:constraints_vec} can then be written as
	\begin{align} \label{eq:constraints_vec2} 
	(\vec{\xi}^\fcalvec)^\Transp \mat{\Phi}_i \mat{\Gamma} \vec{\xi}^\gcalvec = 0,  \qquad \forall \quad i=1,\dots,L.
	\end{align}
	
	We perform the multiplication and collect the terms in $\vec{\xi}^\fcalvec$ and $\vec{\xi}^\gcalvec$. The condition \eqref{eq:constraints_vec2} then results in conditions on the parameters in $\mat{\Gamma}$ resulting a in a homogeneous system of linear equations 	
	\begin{align} \label{eq:parameter_eq}
	\A \cdot \text{vec}(\mat{\Gamma}) = \vec{0}.
	\end{align}
	
	The vectors $\text{vec}(\mat{\Gamma}_1), \dots , \text{vec}(\mat{\Gamma}_P)$ spanning the nullspace of $\A$ in \eqref{eq:parameter_eq} are then used to compute the columns in $\Gcalmat_\x = [\gcalvec_1,\dots \gcalvec_P]$ where $\gcalvec_p = \mat{\Gamma}_p\vec{\xi}^\gcalvec$ .
	If it turns out that the nullspace of $\A$ is empty, one should start over with a new ansatz and extend the set of operators in $\vec{\xi}^\gcalvec$.
	
	The outline of the procedure as described above is summarized in Algorithm \ref{alg:find_G}.
	\begin{algorithm}[tb]
		\caption{Constructing $\Gcalmat_\x$}
		\label{alg:find_G}
		\begin{algorithmic}
			\STATE {\bfseries Input:} Operator matrix $\Fcalmat_\x$
			\STATE {\bfseries Output:} Operator matrix $\Gcalmat_\x$ where $\Fcalmat_\x \Gcalmat_\x = \vec{0}$
			\STATE {\bfseries Step 1:} Make an ansatz  $\gcalvec=\mat{\Gamma}\vec{\xi}^\gcalvec$ for the columns in $\Gcalmat_\x$.
			\STATE  {\bfseries Step 2:} Expand $\Fcalmat_\x\mat{\Gamma}\vec{\xi}^\gcalvec$ 
			and collect terms.
			\STATE  {\bfseries Step 3:} Construct 
			$\A \cdot \text{vec}(\mat{\Gamma}) = \vec{0}$ and find the vectors $\mat{\Gamma}_1 \dots \mat{\Gamma}_P$ spanning its nullspace.
			\STATE {\bfseries Step 4:} If $P=0$, go back to {\bfseries Step 1} and make a new ansatz, i.e. extend the set of operators.
			\STATE {\bfseries Step 5:} Construct $\Gcalmat_\x = [\mat{\Gamma}_1\vec{\xi}^\gcalvec, \dots, \mat{\Gamma}_P\vec{\xi}^\gcalvec]$.
		\end{algorithmic}
	\end{algorithm}
	The algorithm 
	is based upon a parametric ansatz rather than directly upon the theory for linear operators.
	Not only is it more intuitive, but it does also remove any conceptual challenges that theory may provide.
	A problem with this is that one may have to iterate before having found the appropriate set of operators in $\Gcalmat_\x$.
	It might be of interest to examine possible alternatives to this algorithm that does not use a parametric approach.
	Let us now illustrate the method with an example.
	
	\subsection{Divergence-free example revisited}
	Let us return to the example discussed in Section \ref{sec:2d_divfree}, and show how the solution found by visual inspection also can be found with the algorithm described above.
	Since $\Fcalmat_\x$ only contains first-order derivative operators, we assume that a column in $\Gcalmat_\x$ does so as well.
	Hence, let us propose the following ansatz (step 1)
	\begin{align}
	\gcalvec=
	\begin{bmatrix}
	\gamma_{11} & \gamma_{12}\\
	\gamma_{21} & \gamma_{22}
	\end{bmatrix}
	\begin{bmatrix}
	\pd{}{x_1} \\[1mm]
	\pd{}{x_2}
	\end{bmatrix}
	= \Gamma\vec{\xi}^\gcalvec.
	\end{align}
	Applying the constraint, expanding and collecting terms (step 2) we find
	\begin{equation} \label{eq:ex1terms}
	\Fcalmat_\x\Gamma\vec{\xi}^\gcalvec=
	\begin{bmatrix}
	\pd{}{x_1} & \pd{}{x_2}
	\end{bmatrix}
	\begin{bmatrix}
	\gamma_{11} & \gamma_{12}\\
	\gamma_{21} & \gamma_{22}
	\end{bmatrix}
	\begin{bmatrix}
	\pd{}{x_1} \\[1mm]
	\pd{}{x_2}
	\end{bmatrix}
	=	\gamma_{11}\pdd{}{x_1}
	+(\gamma_{12}+\gamma_{21})\pdt{}{x_1}{x_2}
	+\gamma_{22}\pdd{}{x_2},
	\end{equation}
	where we have used the fact that $\pdt{}{x_i}{x_j}=\pdt{}{x_j}{x_i}$ assuming continuous second derivatives.
	The expression \eqref{eq:ex1terms} equals zero if
	\begin{equation}
	\begin{bmatrix}
	1 & 0 & 0 & 0 \\
	0 & 1 & 1 & 0 \\
	0 & 0 & 0 & 1
	\end{bmatrix}
	\begin{bmatrix}
	\gamma_{11} \\
	\gamma_{12} \\
	\gamma_{21} \\
	\gamma_{22}
	\end{bmatrix}
	= \A \cdot \text{vec}(\mat{\Gamma})
	=\vec{0}.
	\end{equation}
	The nullspace is spanned by a single vector (step 3)
	$
	[
	\gamma_{11} \,\,
	\gamma_{12} \,\,
	\gamma_{21} \,\,
	\gamma_{22}
	]^\Transp
	=
	\lambda
	[
	0 \,\, -1 \,\, 1 \,\, 0
	]^\Transp,$  
	$\lambda \in \mathbb{R}.
	$
	Choosing $\lambda=1$, we get
	$
	\Gcalmat_\x=
	\begin{bmatrix}
	-\pd{}{x_2} & \pd{}{x_1}
	\end{bmatrix}^\Transp	
	$
	(step 5), which is the same as in \eqref{eq:Gx1}.
	%

	
	\subsection{Generalization}
	Although there are no conceptual problems with the algorithm introduced above, the procedure of expanding and collecting terms appears a bit informal. 
	In a general form, the algorithm is reformulated such that the operators are completely left out from the solution process.
	The drawback of this is a more cumbersome notation, and we have therefore limited the presentation to this simplified version.
	However, the general algorithm is found in the Supplementary material of this paper.

	\section{Experimental results}
	
	\subsection{Simulated divergence-free function}	
	\label{sec:2ddivexp}	
	Consider the example in Section~\ref{sec:2d_divfree}.		
	An example of a function fulfilling 
	$
	\pd{f_1}{x_1}+\pd{f_2}{x_2}=0
	$	
	is
	\begin{equation}\label{eq:2d_divfree_func}	
	\begin{split}
	f_1(x_1,x_2)&=e^{-ax_1x_2}\big(ax_1\sin(x_1x_2)-x_1\cos(x_1x_2)\big),  \\
	f_2(x_1,x_2)&=e^{-ax_1x_2}\big(x_2\cos(x_1x_2)-ax_2\sin(x_1x_2)\big),
	\end{split}
	\end{equation}
	where $a$ denotes a constant.
	We will now study how the regression of this function differs when using the covariance function found in Section \ref{sec:2d_divfree} as compared to a diagonal covariance function
	$
	\K(\x,\x')=
	k(\x,\x')I.
	$
	The measurements generated are corrupted with Gaussian noise such that
	$
	\y_k=\f(\x_k)+\e_k,
	$
	where
	$
	\e_k \sim \N(\vec{0},\sigma^2I).
	$
	The squared exponential covariance function 	
	$		
	k(\x,\x')=
	\sigma_f^2
	\exp\left[
	{-\frac{1}{2}l^{-2}\|\x-\x'\|^2}
	\right]
	$
	has been used for $k_g$ and $k$ with hyperparameters chosen by maximizing the marginal likelihood.
	We have used the value $a=0.01$ in \eqref{eq:2d_divfree_func}.
	
	%
	
	
	We have used 50 measurements randomly picked over the domain $[0\,\,\,4]\times[0\,\,\,4]$, generated with the noise level $\sigma=10^{-4}$.
	The points for prediction corresponds to a discretization using 20 uniformly distributed points in each direction, and hence a total of $N_P=20^2=400$. 
	We have included the approach described is Section \ref{sec:naiveAppr} for comparison.
	The number of artificial observations have been chosen as random subsets of the prediction points, up to and including the full set.		
	
	The comparison is made with regard to the root mean squared error 
	$
	e_{\text{rms}}=\sqrt{\frac{1}{N_P}\bar{\f}_\Delta^\Transp\bar{\f}_\Delta},
	$
	where  $\bar{\f}_\Delta=\hat{\bar{\f}}-\bar{\f}$
	and 
	$
	\bar{\f}
	$ 
	is a concatenated vector storing the true function values in all prediction points and $\hat{\bar{\f}}$ denotes the reconstructed equivalent. 
	To decrease the impact of randomness, each error value has been formed as an average over 50 reconstructions given different sets of measurements. 
	
	%
	
	
	
\begin{figure}[ht]		
		\begin{center}
			\vspace{0mm}
			\centerline{\includegraphics[width=1.2\textwidth]{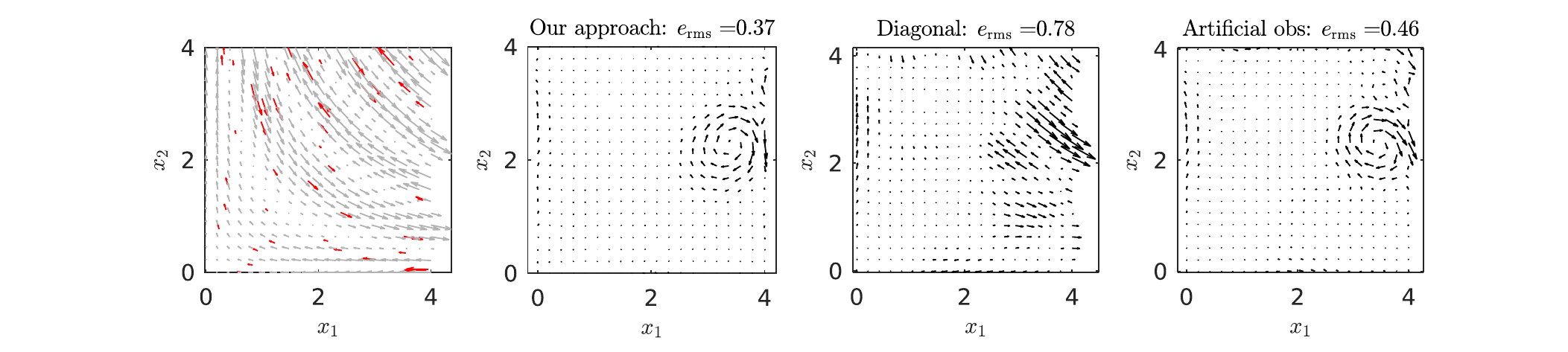}}
			\vspace{-2mm}
			\caption{Left: Example of field plots illustrating the measurements (red arrows) and the true field (gray arrows). Remaining three plots: reconstructed fields subtracted from the true field. 
				The artificial observations of the constraint have been made in the same points as the predictions are made.}
			\label{fig:fieldplot}
		\end{center}
		\vskip -0.3in
	\end{figure}

	An example of the true field, measured values and reconstruction errors using the different methods is seen in Figure \ref{fig:fieldplot}.
	The result from the experiment is seen in Figure \ref{fig:nconst}.
	Note that the error from the approach with artificial observations is decreasing as the number of observations is increased, but only to a certain point.
	Have in mind, however, that the Gram matrix is growing, making the problem larger and worse conditioned.
	The result from our approach is clearly better, while the problem size is kept small and numerical problems are therefore avoided.   
	
	\subsection{Real data experiment}\label{sec:realDataExp}
	Magnetic fields can mathematically be considered as a vector field mapping a 3D position to a 3D magnetic field strength. Based on the magnetostatic equations, this can be modeled as a curl-free vector field. 
	Following Section~\ref{sec:example2sup} in the Supplementary material, our method can be used to encode the constraints in the following covariance function (which also has been presented elsewhere \cite{WahlstromPHD})
	\begin{equation} \label{eq:Kcurl}
	\hspace{-1mm}\K_\text{curl}(\x,\x')=\sigma_f^2e^{{-\frac{\|\x-\x'\|^2}{2l^2}}} 
	\hspace{-1mm}\left(\I_3 \hspace{-0.5mm}-\hspace{-0.5mm}\left(\frac{\x-\x'}{l}\right)\hspace{-1mm}\left(\frac{\x-\x'}{l}\right)^{\hspace{-0.7mm}\Transp}\right).
	\end{equation}
	
	With a magnetic sensor and an optical positioning system, both position and magnetic field data have been collected in a magnetically distorted indoor environment, see the Supplementary material for details about the experimental details. In Figure~\ref{fig:sliceplot} the predicted magnitude of the magnetic field over a two-dimensional domain for three different heights above the floor is displayed. The predictions have been made based on 500 measurements sampled from the trajectory given by the blue curve.
	
	Similar to the simulated experiment in Section~\ref{sec:2ddivexp}, we compare the predictions of the curl-free covariance function \eqref{eq:Kcurl} with the diagonal covariance function and the diagonal covariance function using artificial observations. 	The results have been formed by averaging the error over 50 reconstructions. In each iteration, training data and test data were randomly selected from the data set collected in the experiment. 500 train data points and 1\thinspace000 test data points were used.
	
	
	
\begin{figure}[t]
		\begin{subfigure}[b]{0.5\textwidth}				
			\centerline{\includegraphics[width=\columnwidth]{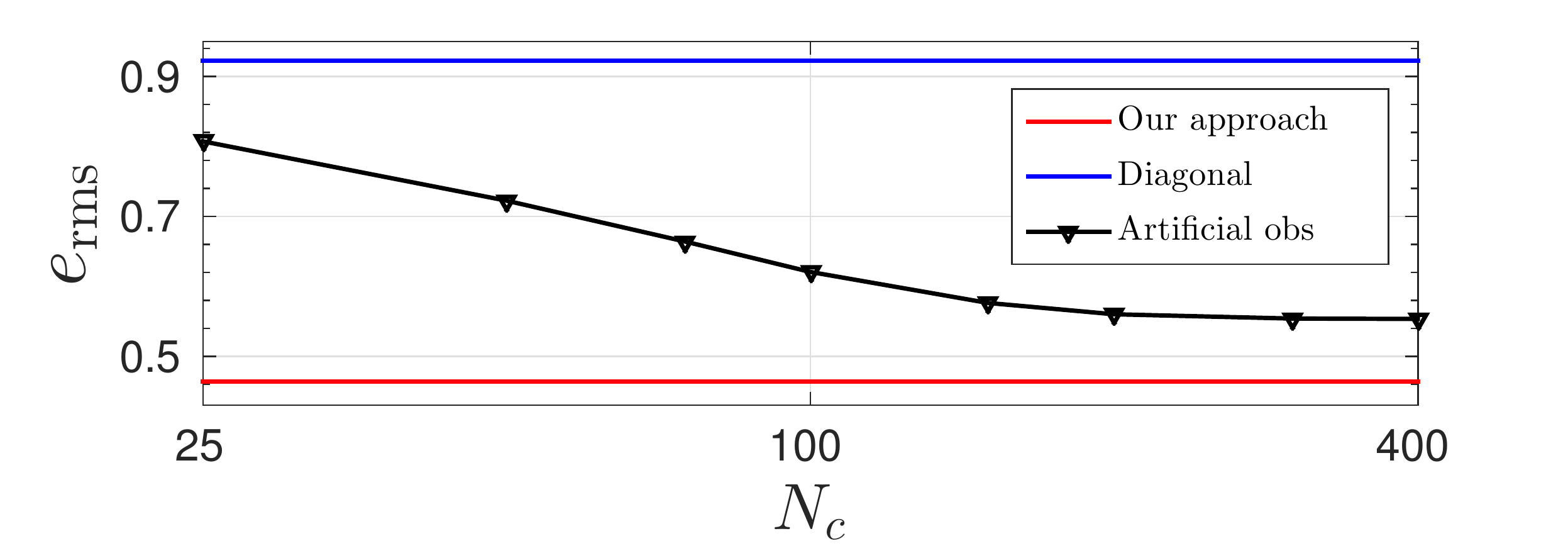}}
			\caption{Simulated experiment}
			\label{fig:nconst}
		\end{subfigure}		
		\begin{subfigure}[b]{0.5\textwidth}							
			\centerline{\includegraphics[width=\columnwidth]{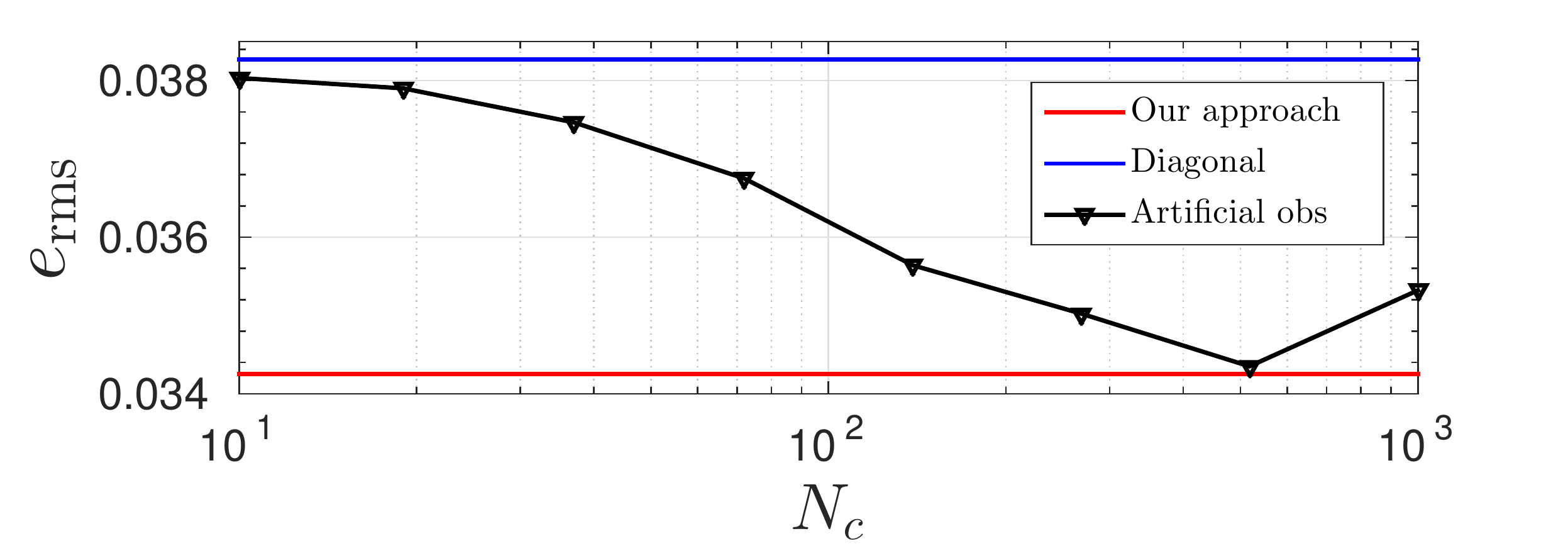}}
			\caption{Real-data experiment}
			\label{fig:ncurl}
		\end{subfigure}					
		\caption{Accuracy of the different approaches as the number of artificial observations $N_c$ is increased.}
		\vspace{-5mm}				
	\end{figure}

	The result is seen in Figure~\ref{fig:ncurl}.
	We recognize the same behavior as we saw for the simulated experiment in Figure~\ref{fig:nconst}.
	Note that the accuracy of the artificial observation approach gets very close to our approach for a large number of artificial observations.
	However, in the last step of increasing the artificial observations, the accuracy decreases. 
	This is probably caused by the numerical errors that follows from an ill-conditioned Gram matrix.
	
	\section{Related work}
	Many problems in which GPs are used contain some kind of constraint that could be well exploited to improve the quality of the solution.
	Since there are a variety of ways in which constraints may appear and take form, there is also a variety of methods to deal with them.
	The treatment of inequality constraints in GP regression have been considered for instance in \cite{Abrahamsen2001} and \cite{DaVeiga2012}, based on local representations in a limited set of points.
	The paper \cite{Maatouk2017} proposes a finite-dimensional GP-approximation to allow for inequality constraints in the entire domain.
	
	It has been shown that linear constraints satisfied by the training data will be satisfied by the GP prediction as well \cite{Salzmann2010ImplicitlyCG}. 
	The same paper shows how this result can be extended to quadratic forms through a parametric reformulation and minimization of the Frobenious norm, with application demonstrated for pose estimation.
	Another approach on capturing human body features is described in \cite{Rudovic_shape-constrainedgaussian}, where a face-shape model is included in the GP framework to imply anatomic correctness.
	A rigorous theoretical analysis of degeneracy and invariance properties of Gaussian random fields is found in \cite{Ginsbourger2017}, including application examples for one-dimensional GP problems.
	
	Although constraints in most situations are formulated on the outputs of the GP, there are also situations in which they are acting on the inputs.
	An example of this is given in \cite{Tran2015}, describing a method of benefit from ordering constraints on the input to reduce the negative impact of input noise.
	
	Applications within medicine include gene-disease association through functional expectation constraints \cite{Koyejo2013} and lung disease sub-type identification using a mixture of GPs and constraints encoded with Markov random fields \cite{ross13a}.
	Another way of viewing constraints is as modified prior distributions.
	By making use of the so-called multivariate generalized 
	von Mises distribution, \cite{Frellsen} ends up in a version of GP regression customized for circular variable problems.
	Other fields of interest include using GPs in approximately solving one-dimensional partial differential equations \cite{Thore2003,Nguyen201569,Nguyen201652}. 
	
	Generally speaking, the papers mentioned above consider problems in which the constraints are dealt with using some kind of external enforcement -- that is, they are not explicitly incorporated into the model, but rely on approximations or finite representations.
	Therefore, the constraints may just be approximately satisfied and not necessarily in a continuous manner, which differs from the method proposed in this paper.
	Of course, comparisons can not be done directly between methods that have been developed for different kinds of constraints.
	The interest in this paper is multivariate problems where the constraints are linear combinations of the outputs that are known to equal zero.  
	
	For multivariate problems, constructing the covariance function is particularly challenging due to the correlation between the output components.
	We refer to \cite{Alvarez2012} for a very useful review. 
	The basic idea behind the so-called \textit{separable kernels} is to separate the process of modeling the covariance function for each component and the process of modeling the correlation between them. 
	The final covariance function is chosen for example according to some method of regularization.
	Another class of covariance functions is the \textit{invariant kernels}.
	Here, the correlation is inherited from a known mathematical relation. 
	The curl- and divergence free covariance functions are such examples where the structure follows directly from the underlying physics, and has been shown to improve the accuracy notably for regression problems  \cite{WahlstromPHD}. 
	Another example is the method proposed in \cite{Constantinescu2013}, where the Taylor expansion is used to construct a covariance model given a known relationship between the outputs. 
	A very useful property on linear transformations is given in \cite{Sarkka2011}, based on the GPs natural inheritance of features imposed by linear operators.
	This fact has for example been used in developing a method for monitoring infectious diseases \cite{Andrade-Pacheco2016}. 
	
	The method proposed in this work is exploiting the transformation property to build a covariance function of the invariant kind for a multivariate GP.
	We show how this property can be exploited to incorporate knowledge of linear constraints into the covariance function.
	Moreover, we present an algorithm of constructing the required transformation.
	This way, the constraints are built into the prior and are guaranteed to be fulfilled in the entire domain.
	
	\section{Conclusion and future work}
	We have presented a method for designing the
	covariance function of a multivariate Gaussian process subject
	to known linear operator constraints on the target function.
	The method will by construction guarantee that any sample
	drawn from the resulting process will obey the constraints in
	all points. Numerical simulations show the benefits of this
	method as compared to alternative approaches.  Furthermore, it
	has been demonstrated to improve the performance on real data
	as well.
	
	As mentioned in Section~\ref{sec:parametric_approach}, it
	would be desirable to describe the requirements on
	$\Gcalmat_\x$ more rigorously. 
	That might
	allow us to reformulate the construction algorithm for~$\Gcalmat_\x$ in a way that allows for a more
	straightforward approach as compared to the parametric ansatz
	that we have proposed. 
	In particular, our method relies upon
	the requirement that the target function can be expressed in
	terms of an underlying \textit{potential} function $\g$. 
	This leads to the intriguing and nontrivial question: Is it
	possible to mathematically guarantee the existence of such a
	potential? 
	If the answer to this question is yes, the next
	question will of course be what it look like and how it relates to the target function. 
	
	
	Another possible topic of further research is the extension to
	constraints including \emph{nonlinear} operators, which for example
	might rely upon a linearization in the domain of interest.
	Furthermore, it may be of potential interest to study the extension to a non-zero right-hand side of~\eqref{eq:constraints}.
	
	\section{Acknowledgements}
	This research is financially supported by the Swedish Foundation for
	Strategic Research (SSF) via the project \emph{ASSEMBLE} (Contract number: RIT 15-0012). 
	The work is also supported by the Swedish Research Council (VR) via the project \emph{Probabilistic modeling of dynamical systems} (Contract number:
	621-2013-5524).
	We are grateful for the help and equipment provided by the UAS Technologies Lab, Artificial Intelligence and Integrated Computer Systems Division (AIICS) at the Department of Computer and Information Science (IDA), Link\"{o}ping University, Sweden.
	The real data set used in this paper has been collected by some of the authors together with Manon Kok, Arno Solin, and Simo S{\"a}rkk{\"a}. 
	We thank them for allowing us to use this data. 
	We also thank Manon Kok for supporting us with the data processing.
	Furthermore, we would like to thank Carl Rasmussen and Marc Deisenroth for fruitful discussions on constrained GPs.
	
	\section{Supplementary material}\label{sec:suppl}
	\subsection{Linear operators}
	\label{app:lin_op}
	In this work we consider linear operators on functions. Such an operator transforms a function $\f(\x)$ to another function $\g(\z)$. We denote this according to
	\begin{align}
	\g(\z) = \Fcalmat_\z[\f(\x)].
	\end{align}
	
	This linear operator could be \emph{differentiation} of a function. If $D=1$ and $K=1$ this will be defined as 
	\begin{subequations}
		\begin{align} \label{eq:diff}
		g(z) & = \Fcal_z[f] = \pd{f(x)}{x}\Big|_{x=z} 
		\end{align}
		which slightly more informal also can be written as	
		\begin{align} \label{eq:int}
		g(x) & = \Fcal_x[f] = \pd{f(x)}{x}. 
		\end{align}
	\end{subequations}
	Also \emph{integration} of a scalar function $f(x)$ over an interval $[z_1,z_2]$  is a linear operator
	\begin{align}  
	g(\z) & = \Fcal_{\z}[f] = \int_{z_1}^{z_2} f(x)dx,
	\end{align}
	where $g(\z)$ is a scalar-valued function with a two-dimensional input $\z=[z_1, \,\,\, z_2]^\Transp$. 	
	Note that in the two examples given above, the inputs of $f$ and $g$ will not be the same, not even of the same dimension! 
	
	Input wrapping is another way to construct new covariance functions from old ones \cite[page 92]{Rasmussen2006}. 
	It utilizes a nonlinear wrapping $\x = \u(\z)$ of the input variables. 
	This wrapping can also be considered as a linear operator, where
	\begin{align}
	\g(\z) & =  \Fcalmat_{\z}[\f]=\f(\x)|_{\x=\u(\z)}.
	\end{align}
	This operator also changes the function input and possibly also its dimension. Even though the wrapping itself might be nonlinear, the operator corresponding to this wrapping is in fact linear. 
	
	It is straightforward to show that all three operators presented above do fulfill the linearity condition.
	
	\subsection{Gaussian processes under linear operations}
	\label{app:gp_lin_op}
	It is well-known that Gaussian distributions are closed under linear transformation. In similar manner, Gaussian processes are closed under linear operations \cite{Papoulis1991,Rasmussen2006,Hennig2013,Garnett2017}. 
	
	By applying the functional $\Fcalmat_\x$ on both the mean function and the covariance function, the GP prior for $\Fcalmat_\x$ is given by			
	\begin{align}
	\Fcalmat_\x \f \sim 
	\GPd{\Fcalmat_\x\ \vec{\mu}}
	{\Cov{\Fcalmat_\x\f(\x)}{\Fcalmat_{\x'}\f(\x')}}.
	\end{align}
	The covariance becomes
	\begin{align}
	\nonumber
	&\Cov{\Fcalmat_\x\f(\x)}{\Fcalmat_{\x'}\f(\x')}
	\\
	\nonumber
	&=
	\mathbb{E}
	\left[
	\big(
	\Fcalmat_\x\f(\x)-\Fcalmat_\x\vec{\mu}(\x)
	\big)
	\big(
	\Fcalmat_{\x'}\f(\x')-\Fcalmat_{\x'}\vec{\mu}(\x')
	\big)^\Transp
	\right]	
	\\
	\nonumber
	&=
	\Fcalmat_\x
	\mathbb{E}
	\left[
	\big(
	\f(\x)-\vec{\mu}(\x)
	\big)
	\big(
	\f(\x')-\vec{\mu}(\x')
	\big)^\Transp
	\right]
	\Fcalmat_{\x'}^\Transp
	\\
	&=
	\Fcalmat_{\x} \K\Fcalmat_{\x'}^\Transp,
	\end{align}
	where by the notation $ (\Fcalmat_\x \K
	\Fcalmat_{\x'}^\Transp)_{ij}$ we mean that
	\begin{align}
	\label{eq:2}
	(\Fcalmat_\x \K \Fcalmat_{\x'}^\Transp)_{ij} 
	&= (\Fcal_\x)_{ik} (\Fcal_{\x'})_{jl} K_{kl},
	\end{align}
	and where
	$(\Fcal_\x)_{ik} $ and $ (\Fcal_{\x'})_{jl}$ act on the first and second argument of $\K_{kl}(\x,\x')$, respectively.
	
	We should point out that some care must be taken when applying this procedure.
	For example, if we would like to consider the derivative of a function governed by a GP, we must make sure that this function is modeled in a way such that the derivative actually exists.
	This may sound obvious, yet important to remember since the set of standard covariance functions includes members that are not differentiable -- among those we find Mat\'{e}rn$_{1/2}$  \cite{Rasmussen2006}.	
	
	\subsection{Generalization of Section~\ref{sec:parametric_approach}}
	
	In this supplementary material we will generalize the method described
	in the main paper on how to solve operator matrix equations on the form
	\begin{align*}
	\Fcalmat \Gcalmat = \vec{0},
	\end{align*}
	where we want to find $\Gcalmat$ given  $\Fcalmat$ \footnote{In this supplementary material, the argument $\x$ is omitted for simplified notation}. If $\Fcalmat \in
	\mathbb{R}^{m \times n}$ is a real valued matrix, $\Gcalmat$ can
	easily be found by letting the columns in $\Gcalmat$ span the
	nullspace of $\Fcalmat$ (provided such a nullspace exist). However, if
	the elements of $\Fcalmat$ are operators, the situation is more
	tricky. This supplementary material generalizes the parametric
	approach presented in Section~\ref{sec:parametric_approach} in the
	main paper for arbitrary operators of any order. The strategy is to study the vector space of homogeneous polynomials where the operators are interpreted as the variables of these polynomials. 
	
	In Section~\ref{sec:first_order}, we assume that both $\Fcalmat$ and $\Gcalmat$ consist of first order operators and in Section~\ref{sec:higher_order} we generalize this to allow for any order of the operators.
	
	\subsubsection{First order operator equation} \label{sec:first_order}
	Consider the matrix $\Fcalmat \in \mathcal{P}_p^{m \times n}$, where $\mathcal{P}_p$ is a vector space of first order operators
	\begin{align}
	\mathcal{P}_p = \{a_1y_1+\dots a_p y_p|a_1,\dots,a_p \in \mathbb{R}\},
	\end{align}
	where $y_1, \dots, y_p$ is the basis in that vector space. The basis components $y_k$ can for example represent derivative operators $y_k=\pd{}{x_k}$. We want to find the vectors $\gcalvec \in \mathcal{P}_p^{n}$ such that $\Fcalmat \gcalvec = \vec{0}$ is fulfilled. We can write $\Fcalmat \in \mathcal{P}_p^{m \times n}$ and $\gcalvec \in \mathcal{P}_p^n$ as
	\begin{subequations}
		\begin{align}
		\Fcal_{ij} & = \sum_{k=1}^p \phi_{ijk}y_k, \qquad 
		\phi_{ijk}=\{\mat{\Phi}\}_{ijk} \in \mathbb{R}, \\
		\gcal_{j} & = \sum_{k=1}^p \gamma_{jk}y_k, \qquad \gamma_{jk} =\{\mat{\Gamma}\}_{jk} \in \mathbb{R},
		\end{align}	
	\end{subequations}
	where $\mat{\Phi} \in \mathbb{R}^{m\times n \times p}$ and $\mat{\Gamma} \in \mathbb{R}^{n\times p}$.
	This gives
	\begin{align}
	\Fcalmat \gcalvec  = \mat{0} \Leftrightarrow
	\sum_{j=1}^n\sum_{k=1}^p\sum_{l=1}^p \phi_{ijk}y_k \gamma_{jl}y_l=0 \quad \forall \,\, i=1:m.
	\end{align}	
	For each $i$, we have a quadratic form		
	\begin{align}
	\y^\Transp \mat{\Phi}_{i}\mat{\Gamma}\y=0, 
	\end{align}
	where $\mat{\Phi}_{i} \in \mathbb{R}^{p \times n}$ with $\{\mat{\Phi}_{i}\}_{kj}=\phi_{ijk}$ and $\mat{\Gamma} \in \mathbb{R}^{n \times p}$ with
	$\{\mat{\Gamma}\}_{jk}=\gamma_{jk}$.
	
	The quadratic form is equal to zero for all $\y$ if and only if
	\begin{align}
	\mat{\Phi}_{i}\mat{\Gamma}+\mat{\Gamma}^\Transp\mat{\Phi}_{i}^\Transp=\mat{0} \quad  \forall \quad i=1:m. 
	\end{align}
	
	
	\subsubsection*{Example 1 (divergence free vector field)}
	We consider the following vector of operators $\Fcalmat \in \mathcal{P}_3^{1 \times 3}$
	\begin{align}
	\Fcalmat = \nabla_\x =		\left[\pd{}{x_1},\,\,\,\pd{}{x_2},\,\,\,\pd{}{x_3}\right],
	\end{align}
	where
	\begin{align}
	\Fcal_{ij} = \sum_{k=1}^3 \phi_{ijk}y_k, \quad \forall
	\quad i=1, \quad j=1, 2, 3,
	\end{align}
	where $y_k = \pd{}{x_k}$. Following the notation introduced above, for this particular operator matrix we have			 			
	\begin{align}
	\mat{\Phi}_1 = 
	\begin{bmatrix}
	1 & 0 & 0 \\
	0 & 1 & 0 \\
	0 & 0 & 1
	\end{bmatrix}.
	\end{align}
	We now want of find a vector $\gcalvec \in \mathcal{P}^{3}$ that fulfills $\Fcalmat \gcalvec = \vec{0}$ for all $\y$. We assume that this operator vector is in $\gcalvec \in \mathcal{P}_3^{3}$ and can be written
	\begin{align}
	\gcal_j = \sum_{k=1}^3 \gamma_{jk} y_k \quad j=1, 2, 3,
	\end{align}
	where $\mat{\Gamma} \in \mathbb{R}^{3 \times 3}$ is unknown. Now we have that
	\begin{subequations}
		\begin{align} \label{eq:ex1mateq}
		& \mat{\Phi}_{1}\mat{\Gamma}+\mat{\Gamma}^\Transp\mat{\Phi}_{1}^\Transp=\mat{0} \\
		& \Rightarrow 				 \begin{bmatrix}
		\gamma_{11} & \gamma_{12}-\gamma_{21} & \gamma_{13}-\gamma_{31} \\
		\gamma_{21}-\gamma_{12} & \gamma_{22} & \gamma_{23}-\gamma_{32} \\
		\gamma_{31}-\gamma_{13} & \gamma_{32}-\gamma_{23} & \gamma_{33}
		\end{bmatrix}=\mat{0},
		\end{align}
	\end{subequations}
	which in turn gives
	\begin{subequations}
		\begin{align}
		\gamma_{11} = 0, \qquad\qquad \gamma_{12}+\gamma_{21} = 0, \\
		\gamma_{22} = 0, \qquad\qquad \gamma_{13}+\gamma_{31} = 0, \\
		\gamma_{33} = 0, \qquad\qquad \gamma_{23}+\gamma_{32} = 0.
		\end{align}
	\end{subequations}
	The nullspace of \eqref{eq:ex1mateq} is then spanned by	
	\begin{align*}
	{\setlength{\arraycolsep}{4pt}
		\mat{\Gamma} = \lambda_1 
		\begin{bmatrix}
		0 & 0 & 0 \\
		0 & 0 & 1 \\
		0 & \text{-}1 & 0
		\end{bmatrix}
		+
		\lambda_2
		\begin{bmatrix}
		0 & 0 & \text{-}1 \\
		0 & 0 & 0 \\
		1 & 0 & 0
		\end{bmatrix}
		+
		\lambda_3 
		\begin{bmatrix}
		0 & 1 & 0 \\
		\text{-}1 & 0 & 0 \\
		0 & 0 & 0
		\end{bmatrix},}
	\end{align*}
	which gives
	\begin{align*}
	\gcalvec = 	 				
	\lambda_1 \hspace{-1mm}				
	\begin{bmatrix}
	0\\
	\pd{}{x_3}\\
	\text{-}\pd{}{x_2}
	\end{bmatrix}
	+
	\lambda_2 \hspace{-1mm}
	\begin{bmatrix}
	\text{-}\pd{}{x_3}\\
	0 \\
	\pd{}{x_1}
	\end{bmatrix}
	+
	\lambda_3 \hspace{-1mm}
	\begin{bmatrix}
	\pd{}{x_2} \\
	\text{-}\pd{}{x_1} \\
	0
	\end{bmatrix}\hspace{-1mm}, 				
	\lambda_1, \lambda_2, \lambda_3 \in \mathbb{R}.
	\end{align*}

	\subsubsection*{Example 2 (curl free vector field)}
	\label{sec:example2sup}
	We consider the following vector of operators $\Fcalmat \in \mathcal{P}_3^{3 \times 3}$
	\begin{align} \label{eq:curlfree}
	\Fcalmat =		
	\begin{bmatrix}
	0 & \pd{}{x_3} & -\pd{}{x_2} \\
	-\pd{}{x_3} & 0 & \pd{}{x_1} \\
	\pd{}{x_2} & -\pd{}{x_1} & 0
	\end{bmatrix},
	\end{align}
	where
	\begin{align}
	\Fcal_{ij} = \sum_{k=1}^3 \phi_{ijk}y_k, \quad \forall \,\, i=1:3, \quad j=1:3,
	\end{align}
	where $y_k = \pd{}{x_k}$. 
	For this particular operator matrix we have			 			
	\begin{align*}
	{\setlength{\arraycolsep}{4pt}
		\mat{\Phi}_1 =
		\begin{bmatrix}
		0 & 0 & 0 \\
		0 & 0 & \text{-}1 \\
		0 & 1 & 0
		\end{bmatrix}\hspace{-1mm}
		,
		\,\,\,
		\mat{\Phi}_2 = 
		\begin{bmatrix}
		0 & 0 & 1 \\
		0 & 0 & 0 \\
		\text{-}1 & 0 & 0
		\end{bmatrix}\hspace{-1mm}
		,
		\,\,\,
		\mat{\Phi}_3 = 
		\begin{bmatrix}
		0 & \text{-}1 & 0 \\
		1 & 0 & 0 \\
		0 & 0 & 0
		\end{bmatrix}}\vspace{-1mm}.
	\end{align*}
	We now want to find a vector $\gcalvec \in \mathcal{P}^{3}$ which fulfills $\Fcalmat \gcalvec = \vec{0}$ for all $\y$. We assume that this operator vector is in $\gcalvec \in \mathcal{P}_3^{3}$ and can be written
	\begin{align}
	\gcal_j = \sum_{k=1}^3 \gamma_{jk} y_k \quad j=1, 2, 3,
	\end{align}
	where $\mat{\Gamma} \in \mathbb{R}^{3 \times 3}$ is unknown. Now we have that
	\begin{align*} \label{eq:ex2mateq}
	\mat{\Phi}_{1}\mat{\Gamma}+\mat{\Gamma}^\Transp\mat{\Phi}_{1}^\Transp=\mat{0} \Rightarrow 				
	{\setlength{\arraycolsep}{4pt} 
		\begin{bmatrix}
		0 		& \text{-}\gamma_{31} 		& \gamma_{21} \\
		\text{-}\gamma_{31} & \text{-}2\gamma_{32} 		& \gamma_{22}\text{-}\gamma_{33} \\
		\gamma_{21} 	& \gamma_{22}\text{-}\gamma_{33} & 2\gamma_{23}
		\end{bmatrix}} & =\mat{0}, \\
	\mat{\Phi}_{2}\mat{\Gamma}+\mat{\Gamma}^\Transp\mat{\Phi}_{2}^\Transp=\mat{0} \Rightarrow 				
	{\setlength{\arraycolsep}{4pt}
		\begin{bmatrix}
		2\gamma_{31} 		& \gamma_{32} 	& \gamma_{33}\text{-}\gamma_{11} \\
		\gamma_{32} 			& 0 		& \text{-}\gamma_{12} \\
		\gamma_{33} \text{-} \gamma_{11} & \text{-}\gamma_{12} 	& \text{-}2\gamma_{13}
		\end{bmatrix}} & =\mat{0}, \\
	\mat{\Phi}_{3}\mat{\Gamma}+\mat{\Gamma}^\Transp\mat{\Phi}_{3}^\Transp=\mat{0} \Rightarrow 				 
	{\setlength{\arraycolsep}{4pt}
		\begin{bmatrix}	
		2\gamma_{21} 		& \gamma_{22}\text{-}\gamma_{11}	& \gamma_{23} \\
		\gamma_{22}\text{-}\gamma_{11}	& \text{-}2\gamma_{12} 		& \text{-}\gamma_{13} \\
		\gamma_{23} 			& \text{-}\gamma_{13} 		& 0
		\end{bmatrix}} & =\mat{0},
	\end{align*}
	which in turn gives
	\begin{subequations}
		\begin{align}
		\gamma_{22}-\gamma_{33} = 0, \qquad \gamma_{23} = 0, \qquad 	\gamma_{32} = 0, \\
		\gamma_{33}-\gamma_{11} = 0, \qquad \gamma_{13} = 0, \qquad 	\gamma_{31} = 0, \\
		\gamma_{22}-\gamma_{11} = 0, \qquad \gamma_{12} = 0, \qquad 	\gamma_{21} = 0.
		\end{align}
	\end{subequations}
	The nullspace of \eqref{eq:ex2mateq} is then spanned by	the single base vector
	\begin{align}
	\mat{\Gamma} = \lambda_1 
	\begin{bmatrix}
	1 & 0 & 0 \\
	0 & 1 & 0 \\
	0 & 0 & 1
	\end{bmatrix}, \quad \lambda_1 \in \mathbb{R},			
	\end{align}
	which gives
	\begin{align} \label{eq:gcurlfree}
	\gcalvec = 	 				
	\lambda_1  				
	\begin{bmatrix}
	\pd{}{x_1}\\
	\pd{}{x_2}\\
	\pd{}{x_3}
	\end{bmatrix}, \quad \lambda_1 \in \mathbb{R}.				
	\end{align}
	The final covariance function becomes
	\begin{equation}\label{eq:curl-free_cov}
	\K(\x,\x') = 
	\left[
	\begin{array}{ccc}
	\frac{\partial^2}{\partial x_1\partial x_1'} & \frac{\partial^2}{\partial x_1\partial x_2'} & \frac{\partial^2}{\partial x_1\partial x_3'} \vspace{2mm} \\ 
	\frac{\partial^2}{\partial x_2\partial x_1'} & \frac{\partial^2}{\partial x_2\partial x_2'} & \frac{\partial^2}{\partial x_2\partial x_3'} \vspace{2mm} \\
	\frac{\partial^2}{\partial x_3\partial x_1'} & \frac{\partial^2}{\partial x_3\partial x_2'} & \frac{\partial^2}{\partial x_3\partial x_3'}              \\
	\end{array}
	\right]k_g(\x,\x').
	\end{equation}
	
	If we use the squared exponential covariance function
	\begin{equation}
	k_g(\x,\x') = \sigma_f^2e^{{-\frac{\|\x-\x'\|^2}{2l^2}}}
	\end{equation}
	we get 
	\begin{equation} \label{eq:Kcurl_supp}
	\hspace{-1mm}\K(\x,\x')=\frac{\sigma_f^2}{l^2}e^{{-\frac{\|\x-\x'\|^2}{2l^2}}} 
	\hspace{-1mm}\left(\I_3 \hspace{-0.5mm}-\hspace{-0.5mm}\left(\frac{\x-\x'}{l}\right)\hspace{-1mm}\left(\frac{\x-\x'}{l}\right)^{\hspace{-0.7mm}\Transp}\right).
	\end{equation}
	This covariance function is used in the real data experiment in Section~\ref{sec:realDataExp} of the main paper. Note, that the version in the paper does not use $l^2$ in the denominator (which we also would get here if we would multiply \eqref{eq:gcurlfree} with $l^2$, still providing the same constraints).

	\begin{figure*}[t]
		\begin{center}
			\includegraphics[width=0.32\linewidth]{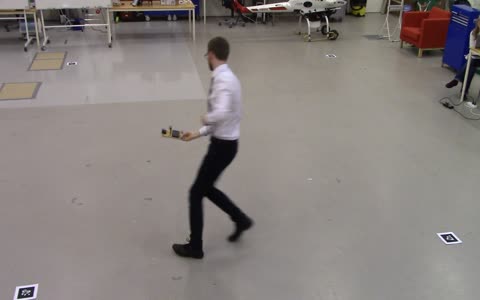}		
			\includegraphics[width=0.32\linewidth]{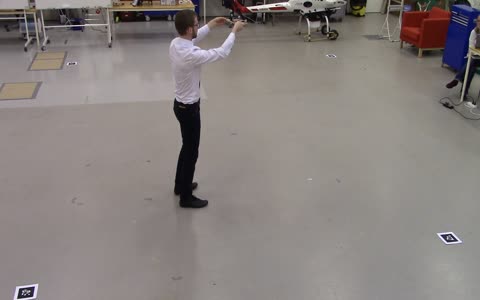}		
			\includegraphics[width=0.32\linewidth]{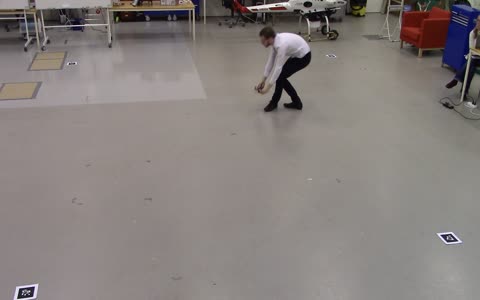}			
			\caption{Three snapshots from the measurement collection. The senor platform was moved around by hand during approximately three minutes.}
			\label{fig:positions}
		\end{center}
	\end{figure*} 		
	
	\subsubsection{Higher order operator equation}	
	\label{sec:higher_order}	
	Now, consider the matrix $\Fcalmat \in \mathcal{P}_{p,q}^{m \times n}$, where $\mathcal{P}_{p,q}$ is a vector space of all homogeneous polynomials of degree $q$ in $p$ variables
	\begin{align*}
	\mathcal{P}_{p,q} = \left\{\sum_{k_1}^{p} \dots \sum_{k_q}^{p} a_{k_1,\dots,k_q} y_{k_1}\cdots y_{k_q} \mat{\Gamma}ig|a_{k_1,\dots,k_q} \in \mathbb{R}\right\},		
	\end{align*}
	where the nominals $y_{k_1}\cdots y_{k_q}$ constitute the basis of that vector space.
	The components $y_k$ can for example represent derivative operators $y_k=\pd{}{x_k}$ and $\mathcal{P}_{p,q}$ then contain all $q$th order derivatives of $x_1 \dots x_q$.
	We want to find the vectors $\gcalvec \in \mathcal{P}_{p,q_g}^{n}$ such that $\Fcalmat \gcalvec = \vec{0}$ is fulfilled. We can write $\Fcalmat \in \mathcal{P}_{p,q}^{m \times n}$ and $\gcalvec \in \mathcal{P}_{p,q_g}^n$ as
	\begin{subequations}
		\begin{align}
		\Fcal_{ij} & = \sum_{k_1}^{p} \dots \sum_{k_q}^{p} \phi_{i,j,k_1,\dots,k_q} y_{k_1}\cdots y_{k_q}, \\
		\gcal_{j} & = \sum_{k_1}^{p} \dots \sum_{k_q}^{p} \gamma_{j,k_1,\dots,k_{q_g}} y_{k_1}\cdots y_{k_{q_g}},
		\end{align}	
	\end{subequations}
	where $\mat{\Phi} \in \mathbb{R}^{m\times n \times p^{\times q}}$ and $\b \in \mathbb{R}^{n\times p^{\times q}}$  (here $p^{\times q}$ denotes $\underbrace{p \times \dots \times p}_{q \text{ times}}$).
	This gives
	\begin{align*}
	& \Fcalmat \gcalvec  = \mat{0} \Leftrightarrow
	\sum_{j}^{n} \sum_{k_1}^{p} \dots \sum_{k_q}^{p} \sum_{l_1}^{p} \dots \sum_{l_q}^{p} \bigg \{ \\
	&  \phi_{ijk_1\dots k_q}y_{k_1}\cdots y_{k_{q}} \gamma_{jl_1\dots l_{q_g}}y_{l_1}\cdots y_{l_{q_g}}\bigg \}=0 \quad \forall \,\, i=1:m.
	\end{align*}	
	For each $i$, this is an algebraic form of order 
	$q+q_g$				
	\begin{align*}				
	\sum_{j}^{n} \sum_{k_1\dots k_q,l_1\dots l_q \in \{d_1 \dots d_{q+q_g}\}} & \phi_{ijd_1\dots d_q} \gamma_{jd_{q+1}\dots d_{q+q_g}}=0 \\
	\forall \quad i=1:m,\quad 
	& k_1 = 1:  p, \quad \dots,\quad  k_q = 1 : p, \notag \\
	& l_1 = 1 : p, \quad \dots, \quad l_q = 1 : p,
	\end{align*}	
	where the second sum sums over all permutations of $k_1\dots k_q,l_1\dots l_q$.

	\subsection{Real data experiment description}
	This section contains more details about the real data experiment described in Section~\ref{sec:realDataExp}. 
	
	\subsubsection{Experiment setup}
\begin{figure}[ht]
		\vskip 0.2in
		\begin{center}
			\centerline{\includegraphics[width=0.7\columnwidth]{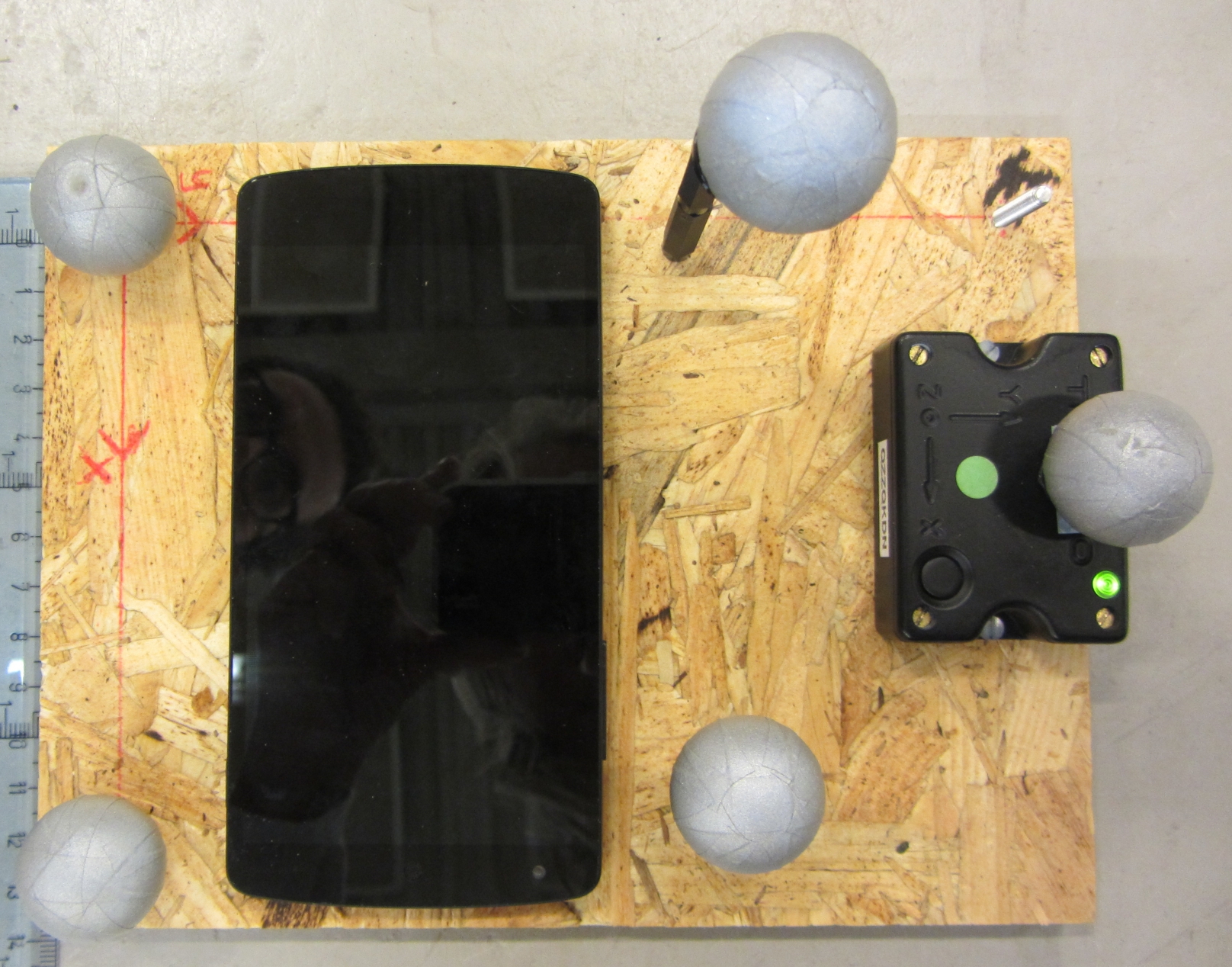}}				
			\caption{Platform with magnetic sensors. The sensor to the left is the Trivisio sensor, whose magnetometer data we used during the experiment. The platform was also equipped with multiple markers visible to an optical reference system (Vicon).}
			\label{fig:platform}
		\end{center}
	\end{figure}

	To collect the measurements we made use of a wooden platform, see Figure~\ref{fig:platform}. The platform was equipped with a Trivisio
	Colibri wireless IMU (TRIVISIO Prototyping GmbH,
	http://www.trivisio.com/), sampled at 100 Hz. The sensor includes both
	an accelerometer, a gyroscope, and a magnetometer. For additional
	validation a Google Nexus 5 smartphone was also mounted on the
	platform even tough its data was never used in this experiment.
	
	On the platform, five markers were mounted. An optical reference system (Vicon) with several cameras mounted in the ceiling measured the 3D position of each marker, and hence also the position and the orientation of the platform relative to its predefined origin.

	\subsubsection{Experiment execution}
	The sensor platform was moved around by hand up and down in a volume of $4 \times 4 \times 2$ meters, see Figure~\ref{fig:positions}. During the experiment, measurements were collected from the sensors on the platform as well from the optical reference system. The data from the different sensors were collected asynchronously. The experiment lasted for 187 seconds.
	
	\subsubsection{Pre-processing of data}
	The position and orientation data from the optical reference system was synchronized with the data from the Trivisio sensor. The synchronization was performed based on correlation analysis of the angular velocities measured by both systems.
	
	The position in global coordinates of the Trivisio sensor was computed based on the position data, the orientation data, and the displacement of the Trivisio sensor relative to the predefined origin of the platform.
	
	The magnetometer data from the Trivisio sensor was rotated from
	sensor-fixed coordinates to global coordinates using the orientation
	data from the optical reference system. These rotated measurements
	describe the magnetic field in global coordinates at the sensor
	positions computed above. In Section~\ref{sec:realDataExp} of the main
	paper, these position data and magnetic field data are considered as input and output data, respectively.	
	
	\bibliography{arxiv}
	\bibliographystyle{plain}
	
\end{document}